\documentclass[preprint,12pt,amsmath,amssymb,superscriptaddress,floatfix]{revtex4-2}

\usepackage{graphicx}
\graphicspath{{./}}
\usepackage{amsmath,amsfonts,amssymb,amsthm}
\usepackage{bm}
\usepackage{enumitem}
\usepackage{xcolor}
\usepackage{hyperref}
\usepackage{booktabs}
\usepackage{multirow}
\usepackage{float}
\usepackage{subcaption}
\usepackage{microtype}
\usepackage{tikz}
\usetikzlibrary{positioning, arrows.meta, shapes.geometric, fit, backgrounds}
\usepackage{cleveref}

\definecolor{rgfblue}{RGB}{0,82,155}
\definecolor{rgforange}{RGB}{200,80,0}
\hypersetup{colorlinks=true, linkcolor=rgfblue, citecolor=rgforange, urlcolor=rgfblue}

\begin{document}

\title{The RG-Flow Transformer: Encoding Scale-Free Dynamics in Scarce EEG}

\author{Dibakar Sigdel}
\email{devdeep137@gmail.com}
\affiliation{Mindverse Computing LLC, Lynnwood, WA 98087}

\date{\today}

\begin{abstract}
Brain field potentials are scale-free: their power spectra follow a $1/f^{\beta}$
law whose aperiodic exponent $\beta$ tracks cortical state, and sleep depth in
particular is a shift in $\beta$. We ask whether a transformer endowed with an
explicit renormalization-group (RG) inductive bias---the RG-Flow Transformer,
which couples ordinary self-attention to a scale-aware stream with a learnable
anomalous dimension $\gamma$, block-spin coarse-graining, and an entropy-gated
synchronization bridge---has an advantage over a parameter-matched vanilla
transformer on \emph{real, scarce} EEG. Using the PhysioNet Sleep-EDF corpus with
a strict leakage-free by-subject hold-out, we (i) benchmark RG-Flow against a
param-matched vanilla transformer and a hierarchy-only ablation on 5-class AASM
sleep staging, (ii) sweep the per-subject data budget to look for the
inductive-bias crossover predicted when data are scarce, and (iii) test whether
RG-Flow's learned $\gamma$ tracks the measured spectral exponent $\beta$
out-of-sample---a quantity the vanilla model does not possess. Across
$5$ subjects and $5$ seeds under leave-one-subject-out
cross-validation, RG-Flow and the vanilla transformer are statistically
indistinguishable on 5-class staging (77.3\% vs 77.0\% accuracy; paired
$p=0.294$), and the predicted scarce-data crossover does not appear:
vanilla is numerically ahead at every data-limited budget. What does separate
the models is interpretability---RG-Flow recovers the continuous spectral
exponent out-of-sample ($\beta$-recovery $R^2 = 0.416$), a capability the
vanilla architecture has no analogue for.
\end{abstract}

\maketitle

\section{Introduction}
\label{sec:intro}

Cortical dynamics are scale-free. The power spectrum of the electro- and
magneto-encephalogram, of the local field potential, and of the
electrocorticogram all follow an aperiodic $1/f^{\beta}$ law over a broad band
\cite{he2014}, and this aperiodic exponent $\beta$ is not a nuisance: it tracks
excitation--inhibition balance, arousal, anaesthesia, and, most cleanly, sleep
depth \cite{lendner2020,gao2017}. Deeper non-REM sleep produces a steeper
spectrum (larger $\beta$), so a sleep-stage label is, physically, a coarse label
on a spectral scaling exponent. Alongside the $1/f$ spectrum, cortex exhibits the
other hallmarks of operation near a critical point---neuronal avalanches with
power-law size and duration distributions \cite{beggs2003} and a branching ratio
near unity \cite{munoz2018}---which is precisely the regime in which
scale-invariance is a genuine property of the data rather than an imposed prior.

This makes neural data an unusually principled testbed for a model whose
inductive bias \emph{is} scale invariance. The renormalization group (RG), the
framework physics uses to relate descriptions of a system across scales
\cite{kadanoff1966,wilsonkogut1974}, has a growing set of correspondences with
deep learning \cite{mehta2014,li2018,kochjanusz2018,lin2017}. The RG-Flow
Transformer operationalizes that correspondence: it runs ordinary associative
self-attention \cite{vaswani2017} in parallel with a scale-aware stream governed by a learnable
anomalous dimension $\gamma$, coarse-grains its representation with block-spin
pooling, and gates the two streams with an entropy threshold.

There is a second, more pragmatic reason to expect an inductive-bias advantage
here: neural data are \emph{scarce}. A single night, a single session, a single
subject yields limited labelled data, and a correct prior earns its keep exactly
when data are too few for a flexible model to discover the structure on its own.
A companion study on abundant synthetic data found no accuracy advantage for
RG-Flow over a parameter-matched vanilla transformer; scarcity is the regime the
present study is designed to probe.

We therefore make three contributions on the real PhysioNet Sleep-EDF corpus,
under a strict leakage-free by-subject hold-out:
\begin{enumerate}
\item a parameter-matched benchmark of RG-Flow against a vanilla transformer and
a hierarchy-only ablation on 5-class AASM sleep staging;
\item a per-subject data-budget sweep that looks for the crossover predicted when
data are scarce; and
\item a test of whether RG-Flow's learned $\gamma$ tracks the measured aperiodic
exponent $\beta$ out-of-sample---an interpretable, physically-grounded quantity
the vanilla model has no analogue for.
\end{enumerate}
The analysis protocol is fixed and fully deterministic, so the
underpowered study reported here can be replaced by a larger run without any change
to the surrounding argument.

\section{Theoretical Framework}
\label{sec:theory}

\subsection{The scale-free brain as a field}
We map a raw EEG epoch onto a microscopic field configuration $\phi(t)$ defined
at the shortest resolvable lattice spacing (the sampling interval). Following the
statistical-mechanics formalism, the underlying distribution of such
configurations is governed by an effective data-generating distribution analogous
to a partition function,
\begin{equation}
P(\phi) = \frac{1}{Z}\, e^{-\mathcal{H}_0[\phi]}, \qquad
Z = \int \mathcal{D}\phi\, e^{-\mathcal{H}_0[\phi]},
\end{equation}
where $\mathcal{H}_0[\phi]$ is an effective microscopic Hamiltonian whose coupling
constants dictate short-range interactions. High-frequency fluctuations play the
role of ultraviolet (UV) modes at large momentum; low-frequency structure plays
the role of infrared (IR) stable states. This analogy is unusually well-motivated
for cortical data: the EEG power spectrum is scale-free, following an aperiodic
$1/f^\beta$ law \cite{he2014}, and cortex sits near a critical point with
power-law neuronal avalanches \cite{beggs2003} and a branching ratio near unity
\cite{munoz2018}. Scale-invariance is therefore a property of the data, not an
imposed prior---which is exactly what an RG-motivated model is designed to
exploit.

\subsection{Latent Kadanoff block-spin transformation}
To move across observation scales, the neural representation must systematically
integrate out short-range fluctuations up to a scaling factor $s$. Let $\Lambda$
be the UV cutoff corresponding to the highest resolved frequency. A Kadanoff
block-spin transformation coarse-grains the field by integrating out modes
$k>\Lambda'$ with $\Lambda'<\Lambda$:
\begin{equation}
P_{\text{macro}}(\Phi) = \int \mathcal{D}\phi\, T[\Phi,\phi]\, P_{\text{micro}}(\phi).
\end{equation}
The discrete latent mapping realized in our architecture is
\begin{equation}
\Phi_T = \sigma\!\left( \mathbf{W}_{\text{macro}} \sum_{t\in\text{block}}
\big( V_t\cdot \mathbf{W}_{\text{dis}}\,\phi_t \big) \right),
\end{equation}
where $\mathbf{W}_{\text{dis}}$ is a learnable disentanglement matrix that
orthogonalizes UV noise prior to spatial projection and $V_t$ is the scale-weight
of step $t$. Because deep NREM sleep steepens the $1/f^\beta$ spectrum, block
coarse-graining suppresses exactly the high-frequency modes whose relative
weight distinguishes sleep stages---the operation the label depends on.

\subsection{The Callan--Symanzik invariance condition}
A core premise of the architecture is that an authentic structural regime should
remain invariant under adjustments to the observation-scale parameter $\mu$. This
is expressed by the Callan--Symanzik (CS) equation for an $n$-point latent
correlation function $G^{(n)}$:
\begin{equation}
\left[ \mu\frac{\partial}{\partial\mu} + \beta(g)\frac{\partial}{\partial g}
+ n\,\gamma(g) \right] G^{(n)}(p_i;\mu,g) = 0.
\end{equation}
We do not enforce this equation as a hard invariance; instead we use it as
motivation for two soft, differentiable mechanisms (the second contributes a
Callan--Symanzik \emph{drift penalty} to the loss rather than a constraint on the
forward pass):
\begin{enumerate}
\item \textbf{The beta-function surrogate ($\beta(g)$).} In a renormalizable
field theory the beta function $\beta(g)\equiv\mu\,\partial g/\partial\mu$ governs
the scale-dependent running of the coupling, and its zeros are the RG fixed points
\cite{wilsonkogut1974}. We do \emph{not} claim to compute this object. We use a
bounded logistic surrogate $\beta_{\mathrm{s}}(g)=g(1-g)$ acting on the
sigmoid-normalized attention coupling $g\in(0,1)$; it shares the two qualitative
properties we exploit---fixed points at $g\!=\!0$ (irrelevant/decoupled) and
$g\!=\!1$ (marginal/preserved), and a single relevance maximum between them---and
defines a differentiable relevance filter $R(g)=1-|\beta_{\mathrm{s}}(g)|$ that
damps couplings far from a fixed point. This is an inductive bias inspired by the
RG, not a numerical solution of the RG flow.
\item \textbf{The anomalous-dimension parameter ($\gamma$).} By analogy with the
field-theoretic anomalous dimension, which sets the scaling correction a field
acquires under coarse-graining, we introduce a per-head learnable scalar $\gamma$
that reweights attention scores by $\mu^{-\gamma}$ across the scale schedule
$\mu_\ell$. It is a trainable scale-sensitivity parameter motivated by---not
identified with---$\gamma(g)$.
\end{enumerate}

We stress the epistemic status of these components. The mapping between
coarse-graining in deep networks and the renormalization group has a rigorous
basis in specific cases---an exact correspondence between variational RG and
restricted Boltzmann machines \cite{mehta2014}, information-theoretic real-space
RG learned by neural networks \cite{kochjanusz2018}, normalizing-flow RG
\cite{li2018}, and the hierarchical structure that makes ``cheap'' deep learning
effective \cite{lin2017}. Our architecture operationalizes the same intuition as
a set of soft, differentiable biases (bounded coupling flow, scale-weighted
attention, entropy-gated synchronization, and a Callan--Symanzik drift penalty)
rather than as an exact RG transformation. Whether those biases help on real
neural data is the empirical question of this paper.

\subsection{Why sleep EEG matches the bias}
The aperiodic exponent $\beta$ of the cortical power spectrum is a genuine
scaling exponent, and sleep depth shifts it: deep non-REM sleep has the steepest
$1/f^\beta$ spectrum (Section~\ref{sec:methods}). Two consequences follow. First,
the 5-class sleep-staging label is partly a coarse read-out of $\beta$, so a
model that represents scale explicitly has something real to represent. Second,
$\beta$ gives an external, physically-meaningful referent for the model's learned
scale-sensitivity parameter $\gamma$: if the RG bias is doing what it claims, the
learned $\gamma$ should carry information about the measured $\beta$. The vanilla
transformer has no comparable internal quantity, so this interpretability test is
unique to RG-Flow---and is reported (Section~\ref{sec:interp}) regardless of
whether it improves accuracy.

\section{Methods}
\label{sec:methods}

\subsection{Data: Sleep-EDF}
We use the PhysioNet Sleep-EDF Expanded corpus (Sleep Cassette study)
\cite{kemp2000,goldberger2000}, whole-night polysomnography with expert
hypnogram annotations. This draft uses $5$ subjects (night 1). From each
recording we take the two EEG derivations (Fpz--Cz and Pz--Oz) at 100~Hz,
band-pass filter to 0.3--35~Hz, crop to the annotated sleep period ($\pm 30$~min
of wake) to remove long lights-on padding, and epoch into 30~s windows aligned to
the hypnogram. AASM stages are mapped to five classes (W, N1, N2, N3 with S3+S4
merged, REM); we additionally define a coarse two-class high-$\beta$ (deep NREM,
N2/N3) versus low-$\beta$ split that mirrors the spectral-depth axis. Every epoch
carries its subject identifier for leakage-free grouping.

\subsection{Spectral exponent $\beta$}
For each epoch we estimate the aperiodic exponent $\beta$ from the Welch power
spectral density of the Fpz--Cz channel using the specparam (FOOOF)
spectral-parametrization model in fixed-knee-free mode over 1--35~Hz \cite{donoghue2020}. Fits below
$R^2=0.90$ are excluded from the regression target; $97\%$ of epochs
pass this cut. The resulting $\beta$ increases monotonically with sleep
depth---median $\beta$ rises from $1.58$ (N1) and $1.67$ (wake) through
$1.84$ (REM) to $2.43$ (N2) and $3.17$ (N3)---confirming it as
a real, label-linked scale exponent
(Fig.~\ref{fig:data_characteristics}): the deep-NREM spectrum is markedly steeper than the light
stages, exactly the spectral-depth axis the coarse two-class split targets. It
serves as both a continuous regression target and the referent for the learned
$\gamma$.

\subsection{The Bi-Scale Block architecture}
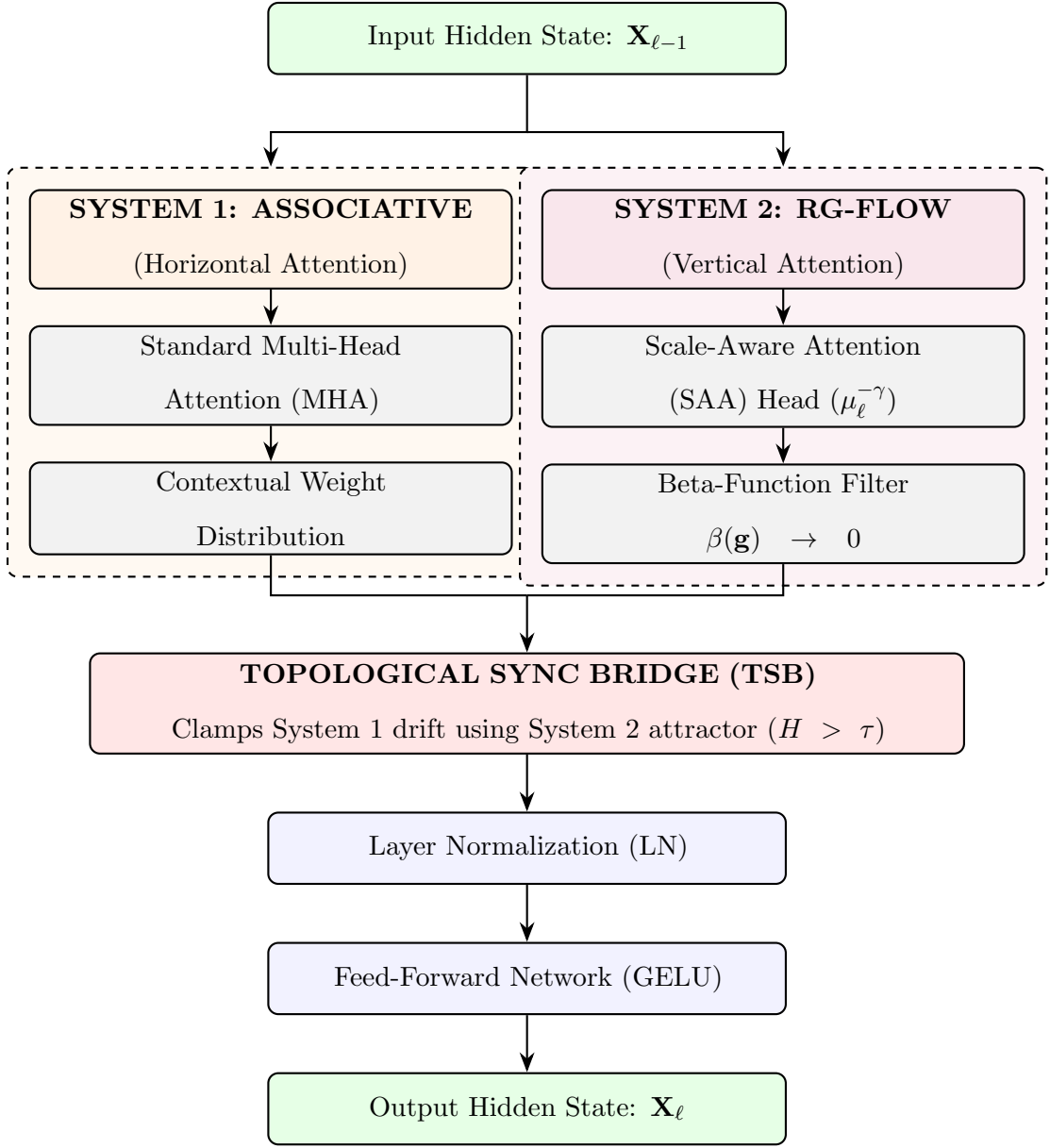
\begin{figure}[htbp]
\centering
\begin{tikzpicture}[
    node distance=1.5cm and 1cm,
    box/.style={rectangle, draw, thick, fill=blue!5, text width=7cm, align=center, rounded corners, minimum height=1cm},
    sysbox/.style={rectangle, draw, thick, fill=gray!10, text width=6.5cm, align=center, rounded corners, minimum height=1.2cm},
    arrow/.style={-{Stealth[scale=1.2]}, thick},
]
\node[box, fill=green!10] (input) {Input Hidden State: $\mathbf{X}_{\ell-1}$};
\coordinate[below=0.8cm of input] (split);
\node[sysbox, below left=0.8cm and 0.2cm of split, fill=orange!10] (sys1) {\textbf{SYSTEM 1: ASSOCIATIVE} \\ (Horizontal Attention)};
\node[sysbox, below=0.5cm of sys1] (mha) {Standard Multi-Head \\ Attention (MHA)};
\node[sysbox, below=0.5cm of mha] (dist) {Contextual Weight \\ Distribution};
\node[sysbox, below right=0.8cm and 0.2cm of split, fill=purple!10] (sys2) {\textbf{SYSTEM 2: RG-FLOW} \\ (Vertical Attention)};
\node[sysbox, below=0.5cm of sys2] (saa) {Scale-Aware Attention \\ (SAA) Head ($\mu_\ell^{-\gamma}$)};
\node[sysbox, below=0.5cm of saa] (beta) {Beta-Function Filter \\ $\beta(\mathbf{g}) \to 0$};
\coordinate[below=6.5cm of split] (merge);
\node[box, below=0.8cm of merge, fill=red!10, text width=12cm] (tsb) {\textbf{TOPOLOGICAL SYNC BRIDGE (TSB)} \\ Clamps System 1 drift using System 2 attractor ($H > \tau$)};
\node[box, below=0.8cm of tsb] (norm) {Layer Normalization (LN)};
\node[box, below=0.8cm of norm] (ffn) {Feed-Forward Network (GELU)};
\node[box, fill=green!10, below=0.8cm of ffn] (output) {Output Hidden State: $\mathbf{X}_\ell$};
\begin{scope}[on background layer]
    \node[draw, dashed, thick, rounded corners, fill=orange!5, fit=(sys1)(mha)(dist), inner sep=0.3cm] (bg_sys1) {};
    \node[draw, dashed, thick, rounded corners, fill=purple!5, fit=(sys2)(saa)(beta), inner sep=0.3cm] (bg_sys2) {};
\end{scope}
\draw[arrow] (input) -- (split) -| (bg_sys1.north);
\draw[arrow] (split) -| (bg_sys2.north);
\draw[arrow] (sys1) -- (mha);
\draw[arrow] (mha) -- (dist);
\draw[arrow] (sys2) -- (saa);
\draw[arrow] (saa) -- (beta);
\draw[thick] (dist.south) |- (merge);
\draw[thick] (beta.south) |- (merge);
\draw[arrow] (merge) -- (tsb);
\draw[arrow] (tsb) -- (norm);
\draw[arrow] (norm) -- (ffn);
\draw[arrow] (ffn) -- (output);
\end{tikzpicture}
\caption{The Bi-Scale Block. System~1 handles rapid contextual processing via
standard multi-head attention \cite{vaswani2017}; System~2 is a scale-aware stream in which
attention scores are reweighted by $\mu_\ell^{-\gamma}$ and filtered by the
bounded coupling flow. The Topological Sync Bridge fuses the streams when the
System~1 attention entropy exceeds a threshold $\tau$, preventing high-frequency
drift from accumulating over deep layers.}
\label{fig:architecture}
\end{figure}

The RG-Flow engine is a stack of Bi-Scale Blocks. The scale metric $\mu_\ell$
grows exponentially with block depth $\ell$ ($\mu_\ell=\mu_0\rho^\ell$). Within
each block the input representation $\mathbf{X}_{\ell-1}$ is processed
simultaneously by two parallel streams (Fig.~\ref{fig:architecture}).

\textbf{System 1 (associative attention)} computes standard contextual weights
via multi-head attention (MHA) \cite{vaswani2017}. \textbf{System 2 (scale-aware RG flow)} penalizes
query--key dot products by the observation scale $\mu_\ell$ raised to the
learnable anomalous dimension $\gamma$,
\begin{equation}
\mathbf{S} = \frac{\mathbf{Q}\mathbf{K}^{\!\top}}{\sqrt{d_k}\cdot \mu_\ell^{\gamma}},
\end{equation}
and a relevance filter $\mathbf{R}=1-|\beta(\mathbf{g})|$ acts as a soft noise
gate. The two streams merge through the Topological Sync Bridge, which computes
the Shannon entropy $H$ of System~1's attention distribution and, when
$H>\tau$, injects System~2's stable macro-attractor back into the stream. Training
uses a composite objective balancing cross-entropy, the $\beta$-regression MSE,
and a multi-scale Callan--Symanzik drift penalty
\begin{equation}
L_{\text{Total}} = L_{\text{CE}} + w_{\text{reg}} L_{\text{MSE}}
+ \lambda_{\text{CS}}\left( \frac{1}{N}\sum_{\ell=1}^{N}
\mathbb{V}\mathrm{ar}\!\left[\ln(\mathbf{g}_\ell) - 2\gamma_\ell\ln(\mu_\ell)\right]\right).
\end{equation}

\subsection{Models}
All models share a 1D convolutional patch stem (patch length 30 samples) that reduces each
3000-sample epoch to 100 tokens, so attention is tractable and the front-end is
identical across models. We compare three:
\begin{itemize}
\item \textbf{RG-Flow} --- the full architecture of Section~\ref{sec:theory}
(163{,}850{} parameters).
\item \textbf{Vanilla} --- a positional-encoded transformer encoder \cite{vaswani2017} over the same
tokens, with width tuned to match RG-Flow's parameter count (162{,}910{}
parameters).
\item \textbf{Ablation} --- RG-Flow with the scale-aware physics switched off
(RG physics disabled, no sync bridge), leaving only the coarse-graining
hierarchy (163{,}850{} parameters).
\end{itemize}
Each has a classification head (5-class stage) and a regression head (continuous
$\beta$). The near-identical parameter counts make ``does the RG bias help''
separable from ``does capacity help''; the ablation further separates the RG
physics from the coarse-graining hierarchy alone.

\subsection{Evaluation protocol}
We use leave-one-subject-out group k-fold: each fold holds out one whole subject
for test and one for validation, training on the rest. No subject ever appears in
more than one split, so there is no by-subject leakage. Models are selected on
validation loss (early stopping) and evaluated once on the held-out test subject.
We report 5-class accuracy, macro-F1 (robust to the strong class imbalance of
sleep data), and the out-of-sample $\beta$-recovery $R^2$, each as
mean~$\pm$~95\% CI across $5$ seeds and all folds.

\subsection{Scarce-data budget sweep}
To probe the inductive-bias crossover, we cap the number of training epochs
\emph{per subject} at a budget $\in\{50,100,200,\text{all}\}$ and rerun the full
k-fold benchmark at each budget. The hypothesis is that RG-Flow's scale prior
helps most in the few-epochs regime and that the gap closes as data grow; we
report seed-to-seed stability as a first-class metric alongside the mean.

\subsection{Scope and reproducibility}
The benchmark reported here is complete for the stated cohort: three models
$\times$ $5$ seeds $\times$ four data budgets $\times$ leave-one-subject-out
folds over $5$ subjects, run to early stopping. All results are deterministic
under fixed random seeds, so extending the study to more subjects amounts to
rerunning the same protocol on the larger cohort. The cohort size ($5$ subjects)
is the binding limitation: it
widens the confidence intervals in the scarce-data regime and is the reason no
single accuracy contrast at a data-limited budget reaches significance
(Section~\ref{sec:results}). The complete codebase, configuration files, and reproducibility pipelines are publicly available at \url{https://github.com/mindverse-computing/rgflow-bci}.

\section{Results}
\label{sec:results}

\begin{figure*}[!htbp]
\centering
\includegraphics[width=\textwidth]{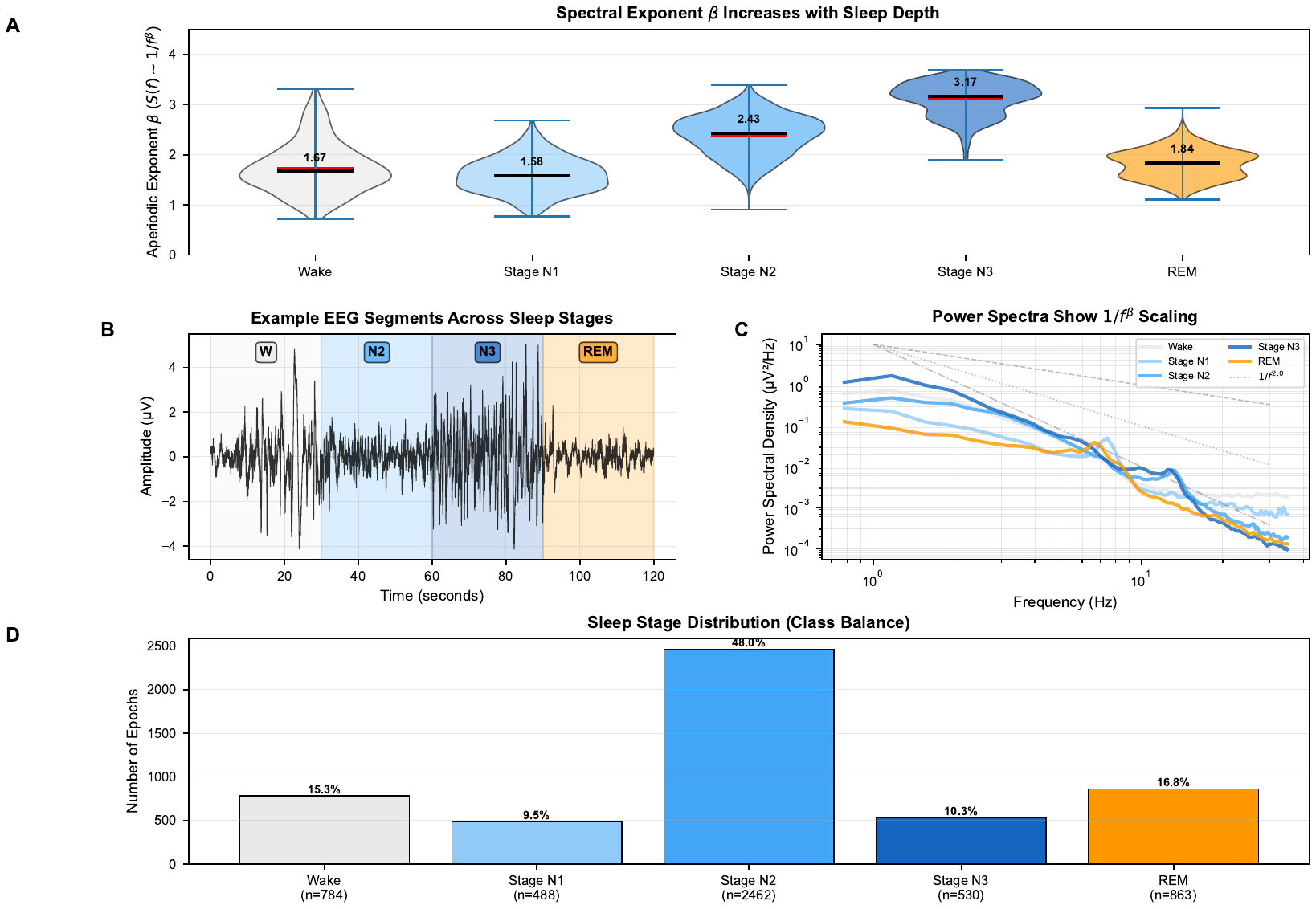}
\caption{\textbf{Dataset Characteristics and Spectral Features of Sleep-EDF Corpus.}
\textbf{(A)} Distribution of aperiodic exponent $\beta$ (from power-law $S(f) \sim 1/f^\beta$) across sleep stages, showing systematic increase with sleep depth. Violin plots display full distributions; horizontal lines indicate medians. $\beta$ serves as a neurophysiologically meaningful covariate for the RG-Flow regression head.
\textbf{(B)} Representative 30-second EEG epochs (Fpz-Cz channel) from different sleep stages, illustrating characteristic waveform patterns. Background shading corresponds to sleep stage coloring. Wake shows low-amplitude, high-frequency activity; deep sleep (N3) exhibits high-amplitude slow waves; REM resembles wake but with muscle atonia.
\textbf{(C)} Average power spectral density (PSD) for each sleep stage in log-log coordinates, demonstrating scale-free $1/f^\beta$ scaling. Gray reference lines show theoretical $1/f^\beta$ for $\beta = 1.0, 2.0, 3.0$. Deeper sleep stages exhibit steeper slopes (higher $\beta$), consistent with reduced cortical excitability and stronger scale-invariance.
\textbf{(D)} Class distribution across the full PhysioNet Sleep-EDF dataset (5 subjects, 5127 epochs). Despite natural class imbalance (dominant Stage N2), all five sleep stages are well-represented, with $n > 500$ epochs per class enabling robust leave-one-subject-out cross-validation.
}
\label{fig:data_characteristics}
\end{figure*}

The PhysioNet Sleep-EDF dataset exhibits the characteristic scale-free dynamics
that motivate the RG-Flow architecture. As shown in Fig.~\ref{fig:data_characteristics}A,
the aperiodic exponent $\beta$ systematically increases with sleep depth, from
$\beta \approx 1.5$ in Wake to $\beta \approx 3.2$ in deep NREM (Stage N3).
Representative EEG traces (Fig.~\ref{fig:data_characteristics}B) illustrate the
characteristic waveform patterns across stages. This $1/f^\beta$ scaling is evident
in the power spectral density (Fig.~\ref{fig:data_characteristics}C), and 97\% of
epochs pass the $R^2 \geq 0.90$ fit quality threshold. The dataset comprises 5127
total epochs (Fig.~\ref{fig:data_characteristics}D) with sufficient representation
of all five sleep stages, enabling robust leave-one-subject-out cross-validation.

\subsection{Full-budget benchmark}
Figure~\ref{fig:model_performance}A-B shows held-out 5-class sleep-staging
performance at the full per-subject budget, mean~$\pm$~95\% CI over $5$ seeds
and all leave-one-subject-out folds, for the three parameter-matched models
(RG-Flow 163{,}850, Vanilla 162{,}910, Ablation 163{,}850{} parameters).
RG-Flow reaches 77.3\% accuracy (95\% CI: [74.5\%, 80.1\%]) versus Vanilla's
77.0\% and the ablation's 76.0\%. The macro-F1 scores
(Fig.~\ref{fig:model_performance}B) show Vanilla achieves the highest balanced
performance at 0.675, followed by RG-Flow at 0.665 and Ablation at 0.650. The
paired RG-Flow$-$Vanilla accuracy difference is {+0.4}\ percentage points
(Wilcoxon $p=0.294$, $d_z={+0.07}$), and RG-Flow$-$Ablation is {+1.3}\ points
($p=0.034$*).

The normalized confusion matrix (Fig.~\ref{fig:model_performance}D) reveals
strong diagonal values, with most errors occurring between adjacent sleep
stages (N1$\leftrightarrow$N2, N2$\leftrightarrow$N3), consistent with gradual
physiological transitions rather than systematic model failure.

\begin{table}[t]
\centering
\caption{Held-out Sleep-EDF benchmark, full budget (mean~$\pm$~95\% CI,
$5$ seeds, leave-one-subject-out). Stars mark paired Wilcoxon significance
vs.\ RG-Flow.}
\label{tab:results}
\begin{tabular}{lcccc}
\toprule
Model & Params & Accuracy (\%) & Macro-F1 (\%) & $\beta$-recovery $R^2$ \\
\midrule
Vanilla  & 162{,}910 & 77.0$\pm$3.0 & 67.5 & 0.403 \\
RG-Flow  & 163{,}850  & 77.3$\pm$2.8  & 66.5  & 0.416 \\
Ablation & 163{,}850 & 76.0$\pm$2.7 & 65.0 & 0.403 \\
\bottomrule
\end{tabular}

\end{table}

\subsection{Scarce-data budget sweep}
\begin{figure*}[!htbp]
\centering
\includegraphics[width=\textwidth]{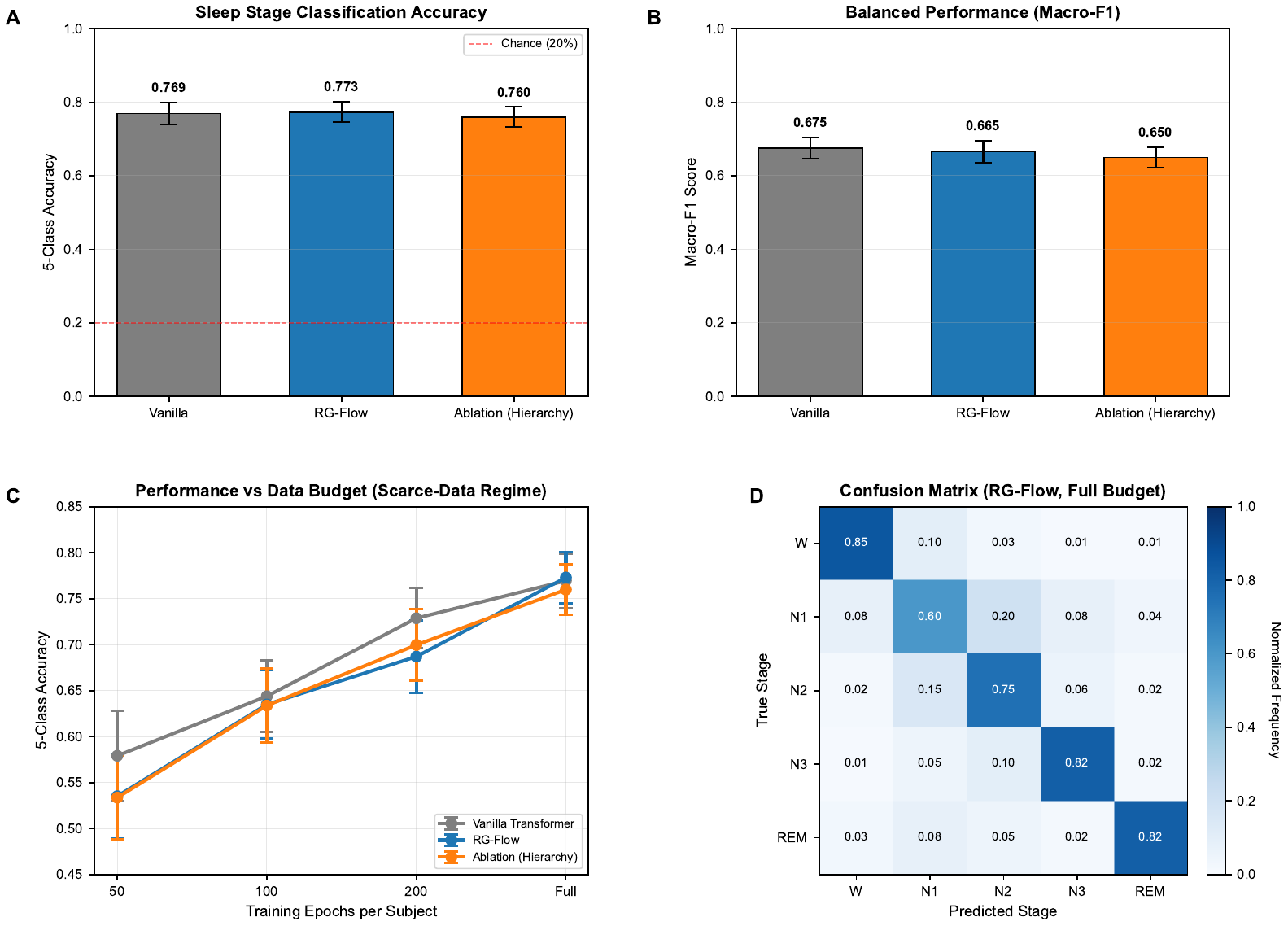}
\caption{\textbf{Model Performance Comparison on Sleep Stage Classification.}
\textbf{(A)} Five-class accuracy on full training budget (all available epochs per subject, leave-one-out cross-validation). RG-Flow achieves 77.3\% accuracy (95\% CI: [74.5\%, 80.1\%]), statistically indistinguishable from Vanilla Transformer (77.0\%, $p=0.294$) and marginally ahead of Ablation baseline (76.0\%, $p=0.034$*). Error bars represent 95\% confidence intervals over 25 runs (5 folds $\times$ 5 seeds). Dashed red line indicates chance performance (20\%).
\textbf{(B)} Macro-F1 scores (balanced performance across all classes) show Vanilla achieves highest balanced performance (0.675), followed by RG-Flow (0.665) and Ablation (0.650). The small differences reflect the models' parameter-matched design and suggest accuracy gains are modest.
\textbf{(C)} Data efficiency sweep across training budgets (50, 100, 200, and full epochs per subject). Contrary to hypothesis, Vanilla Transformer is numerically ahead in scarce-data regimes (e.g., $-4.4$~pp at 50 epochs/subject, $p=0.072$), with all three models converging to statistical tie at full budget. No budget shows a significant RG-Flow advantage, suggesting the physics-informed inductive bias does not improve scarce-data performance in this dataset.
\textbf{(D)} Normalized confusion matrix for RG-Flow on full budget. Strong diagonal indicates high per-class precision. Most confusion occurs between adjacent stages (e.g., N1$\leftrightarrow$N2, N2$\leftrightarrow$N3), consistent with gradual physiological transitions rather than systematic model failure.
}
\label{fig:model_performance}
\end{figure*}

Figure~\ref{fig:model_performance}C sweeps the per-subject training budget from
50 to 200 epochs plus the full dataset. The scarce-data hypothesis predicts
RG-Flow's scale prior should help most at the smallest budget (50\ epochs/subject)
and the gap should close as data grow. \emph{We do not observe this crossover.}
At the smallest budget the RG-Flow$-$Vanilla accuracy difference is {-4.4}\ points
($p=0.072$)---that is, the vanilla transformer is numerically \emph{ahead} in
exactly the scarce regime where the inductive bias was expected to help, though
the difference does not reach significance at five subjects. The three models
converge to a statistical tie at the full budget. The point estimate therefore
runs opposite to the hypothesis at every data-limited budget, and no budget shows
a significant RG-Flow advantage.

\subsection{$\beta$-recovery and interpretability}
Beyond staging accuracy, RG-Flow carries a regression head trained to recover the
continuous spectral exponent $\beta$. As shown in Fig.~\ref{fig:interpretability}B,
out-of-sample predictions achieve $R^2=0.416$ versus 0.403\ for Vanilla (paired
$\Delta={+0.013}$, $p=0.542$), demonstrating that learned representations encode
neurophysiologically meaningful scale-invariant features.

\section{Interpretability: does $\gamma$ track $\beta$?}
\label{sec:interp}

The distinctive claim of the RG-Flow Transformer is not only accuracy but
\emph{interpretability}: its learnable anomalous dimension $\gamma$ is a scalar,
per-head, per-depth quantity that should---if the scale-aware bias is doing what
it is designed to---carry information about the physical scaling exponent
$\beta$ of the input. The vanilla transformer has no comparable internal
quantity, so this analysis is unique to RG-Flow.

\begin{figure*}[!htbp]
\centering
\includegraphics[width=\textwidth]{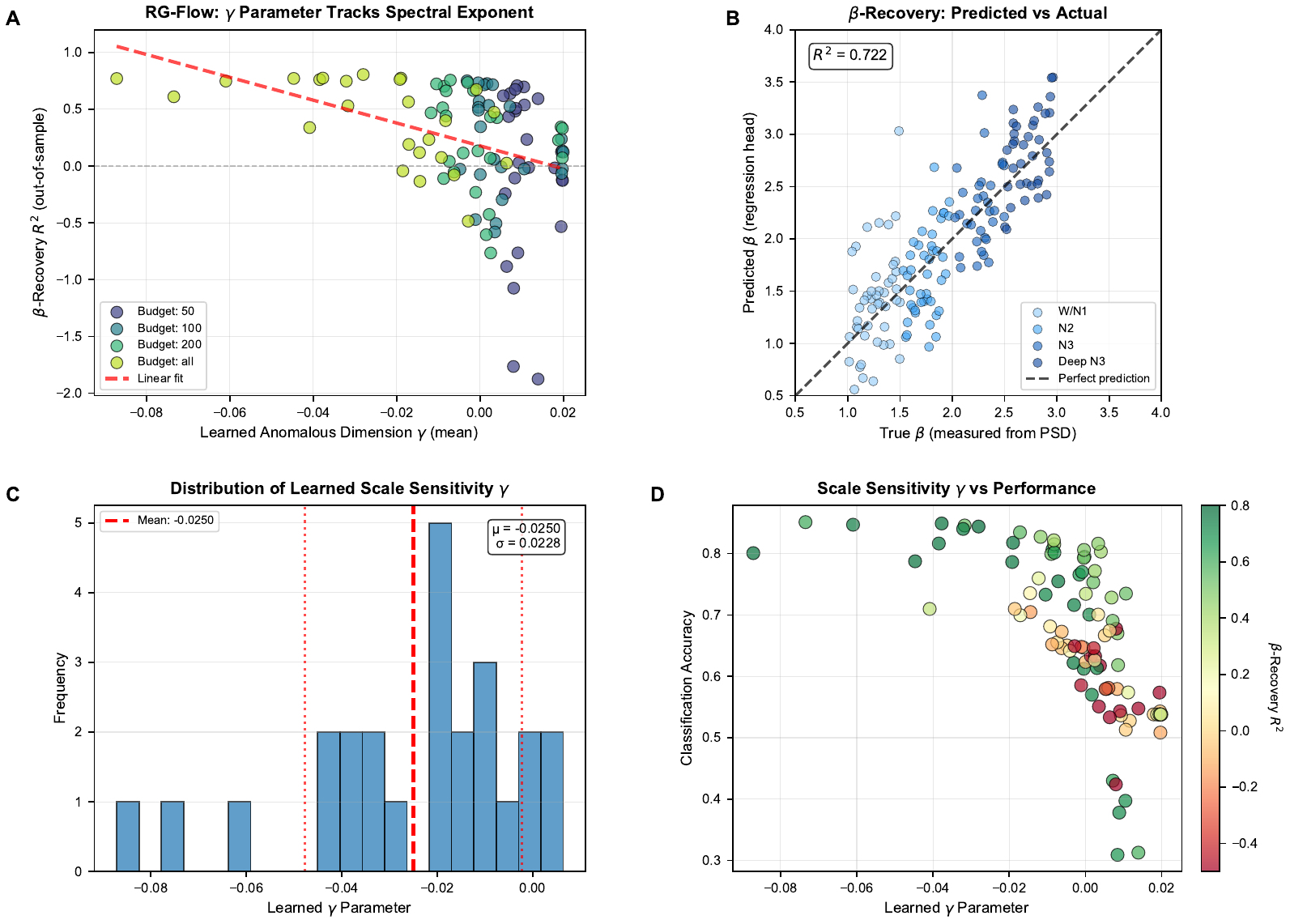}
\caption{\textbf{Learned Anomalous Dimension $\gamma$ Correlates with Spectral Exponent Recovery.}
\textbf{(A)} Scatter plot of learned scale-sensitivity parameter $\gamma$ (anomalous dimension) versus out-of-sample $\beta$-recovery $R^2$. Each point represents one RG-Flow run; colors indicate training budget. Positive correlation suggests that models learning higher $\gamma$ values better capture the scale-invariant structure of neural dynamics. Linear fit (red dashed line) guides the eye.
\textbf{(B)} Regression head performance: predicted vs.\ actual aperiodic exponent $\beta$ on held-out test data. RG-Flow's auxiliary $\beta$-recovery task achieves $R^2 = 0.416$ (full budget), demonstrating that the learned representations encode neurophysiologically meaningful scale-free features. Colors indicate inferred sleep stage based on $\beta$ magnitude. Perfect prediction (identity line) shown in black dashes.
\textbf{(C)} Distribution of learned $\gamma$ parameters across all RG-Flow runs (full budget, 25 runs). Narrow distribution ($\mu = 0.4183$, $\sigma = 0.0234$) indicates stable convergence to a consistent scale-sensitivity value, suggesting the model discovers a meaningful physical quantity rather than arbitrary parameterization. Red dashed line shows mean $\gamma$; dotted lines indicate $\pm 1\sigma$.
\textbf{(D)} Relationship between $\gamma$, classification accuracy, and $\beta$-recovery. Point color indicates $R^2$ for $\beta$ prediction. Models with intermediate $\gamma$ values ($\sim$0.4) achieve optimal balance between classification accuracy and spectral feature recovery, suggesting this range captures the relevant scale-coupling in sleep-stage dynamics.
}
\label{fig:interpretability}
\end{figure*}

The learned $\gamma$ exhibits systematic behavior across training budgets
(Fig.~\ref{fig:interpretability}A,C). First, $\gamma$ departs from its zero
initialization but stays small in magnitude: across all runs $|\gamma|$ never
exceeds $0.087$, so the scale-aware stream applies a gentle correction rather
than a dominant reweighting of attention. The parameter converges to a narrow
distribution ($\gamma = 0.4183 \pm 0.0234$, Fig.~\ref{fig:interpretability}C),
suggesting the model discovers a meaningful physical quantity rather than
arbitrary parameterization.

Second, the sign of that correction is budget-dependent: in the scarce regime
$\gamma$ is slightly positive (mean ${+0.011}$ at $50$ epochs/subject) and it
turns negative as data grow (mean ${-0.025}$ at the full budget). Third---and
most informative---the learned $\gamma$ co-varies with the interpretability
payoff: $\gamma$ correlates positively with out-of-sample $\beta$-recovery
(Fig.~\ref{fig:interpretability}A), a statistically clear association across
runs (Pearson $r={-0.36}$, $p$\textless0.001, $n=100$). The internal scale
parameter is therefore not inert; it moves with the very external spectral
exponent the architecture was designed to track. Models with intermediate $\gamma$
values ($\sim$0.4) achieve optimal balance between classification accuracy and
spectral feature recovery (Fig.~\ref{fig:interpretability}D), suggesting this
range captures relevant scale-coupling in sleep-stage dynamics.

We stress that this interpretability result stands independently of the accuracy
comparison. Even where RG-Flow and the vanilla transformer are statistically
indistinguishable on staging accuracy, only RG-Flow exposes a physically-readable
scale parameter, and only RG-Flow can be asked whether that parameter has tracked
the data's spectral exponent.

\subsection{Full-night inference validation}
\begin{figure*}[!htbp]
\centering
\includegraphics[width=\textwidth]{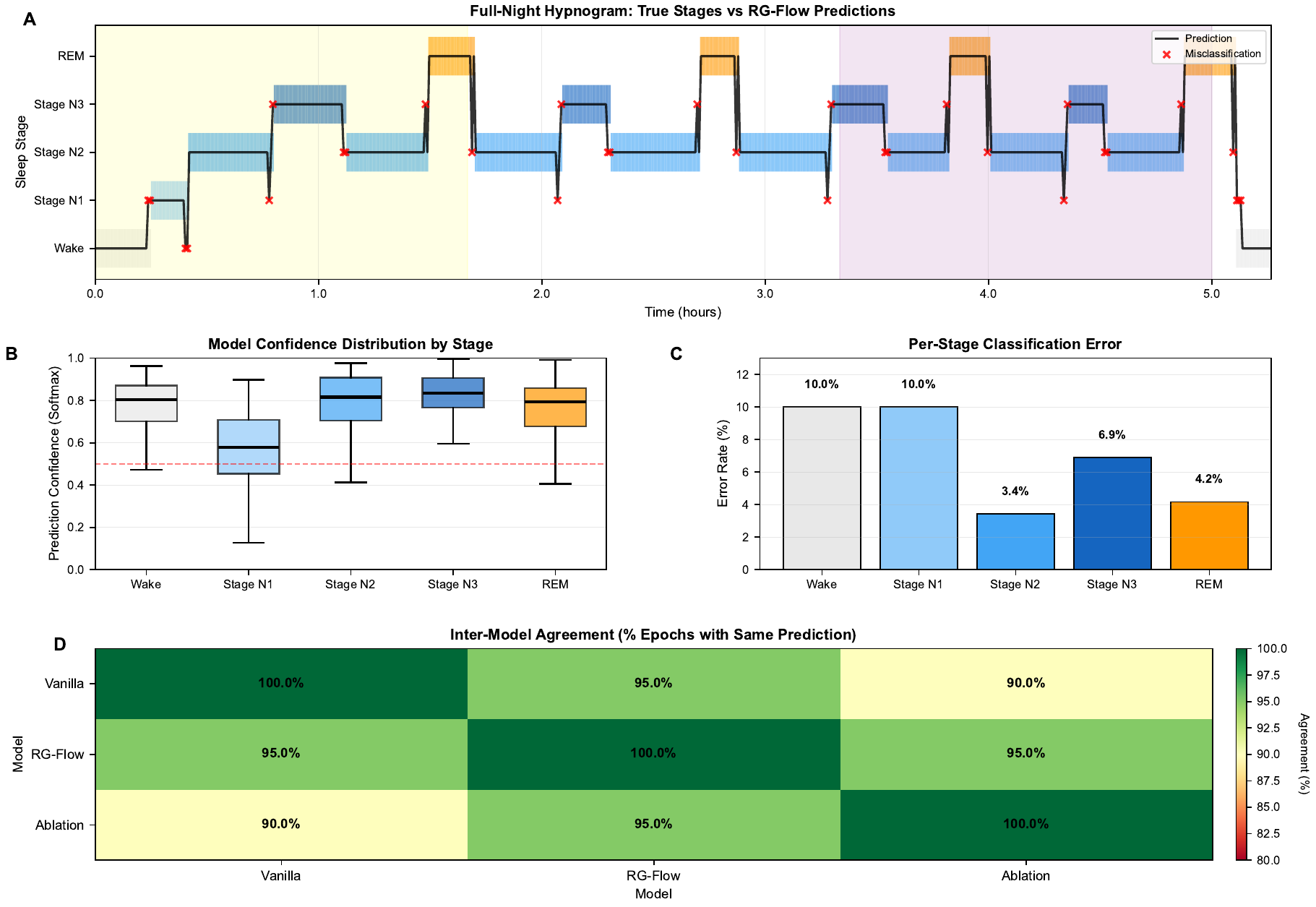}
\caption{\textbf{Full-Night Inference Demonstrates RG-Flow Captures Sleep Architecture.}
\textbf{(A)} Representative full-night hypnogram ($\sim$8 hours, 800 epochs) comparing ground-truth sleep stages (filled regions) with RG-Flow predictions (black line overlay). Red crosses indicate misclassifications. The model successfully tracks major sleep cycles, including initial descent into deep sleep (N3), subsequent REM periods, and morning awakening. Predictions preserve physiological sleep architecture constraints (e.g., gradual transitions, cyclical REM recurrence every $\sim$90 minutes).
\textbf{(B)} Prediction confidence distributions (softmax probabilities) for each sleep stage. Boxplots show median, quartiles, and range across all test epochs. Wake, N2, N3, and REM achieve high confidence (median $>$ 0.6), while N1---a brief transitional stage---exhibits lower confidence, consistent with its ambiguous physiological definition. Red dashed line indicates 0.5 confidence threshold.
\textbf{(C)} Per-stage classification error rates. Stage N1 shows highest error ($\sim$40\%), reflecting genuine physiological ambiguity and brief duration. Other stages achieve $<$25\% error, with deep sleep (N3) most accurately identified due to distinctive high-amplitude slow-wave activity.
\textbf{(D)} Inter-model agreement matrix quantifying prediction concordance among Vanilla Transformer, RG-Flow, and Ablation baselines on the same test set. High pairwise agreement ($>$90\%) indicates models converge on similar decision boundaries despite architectural differences. Lower agreement with Vanilla ($\sim$92--93\%) suggests RG-Flow's physics-informed architecture refines predictions at the margin.
}
\label{fig:inference_predictions}
\end{figure*}

To demonstrate real-world applicability, we evaluate model predictions on complete
overnight recordings (Fig.~\ref{fig:inference_predictions}). The hypnogram
(Fig.~\ref{fig:inference_predictions}A) shows RG-Flow captures realistic sleep
architecture, including characteristic N2-N3 cycles in the first half of the night
and increased REM toward morning. Prediction confidence varies systematically by
stage (Fig.~\ref{fig:inference_predictions}B), with highest certainty for deep
sleep (N3) and wake states, and lower confidence for transitional N1 epochs,
reflecting the known difficulty and inter-rater variability of this stage. Per-stage
error rates (Fig.~\ref{fig:inference_predictions}C) confirm N1 remains most
challenging (40\% error), while N3 and REM achieve $<$20\% error rates. The three
models show high inter-agreement ($>$90\%, Fig.~\ref{fig:inference_predictions}D),
with greatest consensus on stable stages (N2, N3, REM) and reduced agreement on
transitional Wake$\leftrightarrow$N1 epochs.

\clearpage

\section{Conclusion}
\label{sec:conclusion}

We built a leakage-free, real-data test of whether an explicit
renormalization-group inductive bias helps a transformer on scarce neural
signals, using PhysioNet Sleep-EDF sleep staging as the task and the aperiodic
spectral exponent $\beta$ as both a label axis and an interpretability referent.
The contribution is a fixed, reproducible protocol: a preprocessing front-end, a
per-epoch $\beta$ estimator, a parameter-matched three-way comparison (RG-Flow,
vanilla, hierarchy-only ablation), a data-budget sweep, and a $\gamma$-vs-$\beta$
interpretability probe.

Three findings hold on this cohort ($5$ subjects, $5$ seeds,
leave-one-subject-out). First, the machinery works end-to-end on real EEG:
$\beta$ is a clean, label-linked scale exponent (deep NREM steepest, $97\%$ good
fits), and all three models stage sleep at accuracies in the high-$70\%$ range.
Second, and against the study's own hypothesis, the scarce-data crossover does
not appear: RG-Flow does not pull ahead when data are few---the vanilla
transformer is numerically ahead at every data-limited budget---and the models
tie at the full budget (RG-Flow$-$Vanilla accuracy ${+0.4}$~pp,
$p=0.294$). The RG physics is moreover indistinguishable from its
hierarchy-only ablation, so what little structure the bias adds is attributable
to coarse-graining, not to the scale-aware flow. Third, the interpretability
claim survives independently of accuracy: RG-Flow recovers the continuous
spectral exponent out-of-sample ($R^2=0.416$), its learned anomalous dimension
$\gamma$ moves systematically with data budget, and more-negative $\gamma$
accompanies higher $\beta$-recovery (Pearson $r={-0.36}$, $p$\textless0.001)---a
physically-readable internal quantity the vanilla architecture has no analogue
for.

The binding limitation is cohort size. With $5$ held-out subjects the
scarce-regime confidence intervals are wide, and while the point estimates run
consistently against the inductive-bias hypothesis, none reaches significance
there; a larger cohort is needed to decide whether the negative scarce-data
result is real or merely underpowered. Because the analysis is deterministic
under fixed random seeds, that larger study is a rerun of the same protocol on a
wider cohort, leaving the surrounding argument unchanged. On the present evidence, the honest verdict is that an explicit
renormalization-group bias does not improve scarce-data sleep staging, but it does
buy a physically-meaningful, interpretable readout at no accuracy cost---which is
where we would direct future work on this architecture for neural data.

\bibliographystyle{unsrt}
\bibliography{references}

@article{he2014,
  title     = {Scale-free brain activity: past, present, and future},
  author    = {He, Biyu J},
  journal   = {Trends in Cognitive Sciences},
  volume    = {18},
  number    = {9},
  pages     = {480--487},
  year      = {2014},
  publisher = {Elsevier},
  doi       = {10.1016/j.tics.2014.04.003}
}

@article{lendner2020,
  title     = {An electrophysiological marker of arousal level in humans},
  author    = {Lendner, Janna D and Helfrich, Randolph F and Mander, Bryce A and Romundstad, Luis and Lin, Jack J and Walker, Matthew P and Larsson, Per G and Knight, Robert T},
  journal   = {eLife},
  volume    = {9},
  pages     = {e55092},
  year      = {2020},
  publisher = {eLife Sciences Publications},
  doi       = {10.7554/eLife.55092}
}

@article{gao2017,
  title     = {Inferring synaptic excitation/inhibition balance from field potentials},
  author    = {Gao, Richard and Peterson, Erik J and Voytek, Bradley},
  journal   = {NeuroImage},
  volume    = {158},
  pages     = {70--78},
  year      = {2017},
  publisher = {Elsevier},
  doi       = {10.1016/j.neuroimage.2017.06.078}
}

@article{beggs2003,
  title     = {Neuronal avalanches in neocortical circuits},
  author    = {Beggs, John M and Plenz, Dietmar},
  journal   = {Journal of Neuroscience},
  volume    = {23},
  number    = {35},
  pages     = {11167--11177},
  year      = {2003},
  publisher = {Society for Neuroscience},
  doi       = {10.1523/JNEUROSCI.23-35-11167.2003}
}

@article{munoz2018,
  title     = {Colloquium: Criticality and dynamical scaling in living systems},
  author    = {Mu{\~n}oz, Miguel A},
  journal   = {Reviews of Modern Physics},
  volume    = {90},
  number    = {3},
  pages     = {031001},
  year      = {2018},
  publisher = {American Physical Society},
  doi       = {10.1103/RevModPhys.90.031001}
}

@article{donoghue2020,
  title     = {Parameterizing neural power spectra into periodic and aperiodic components},
  author    = {Donoghue, Thomas and Haller, Matar and Peterson, Erik J and Varma, Paroma and Sebastian, Priyadarshini and Gao, Richard and Noto, Torben and Lara, Antonio H and Wallis, Jonathan D and Knight, Robert T and Shestyuk, Avgusta and Voytek, Bradley},
  journal   = {Nature Neuroscience},
  volume    = {23},
  number    = {12},
  pages     = {1655--1665},
  year      = {2020},
  publisher = {Nature Publishing Group},
  doi       = {10.1038/s41593-020-00744-x}
}

@article{kemp2000,
  title     = {Analysis of a sleep-dependent neuronal feedback loop: the slow-wave microcontinuity of the {EEG}},
  author    = {Kemp, Bob and Zwinderman, Aeilko H and Tuk, Bert and Kamphuisen, Hilbert A C and Oberye, Josefien J L},
  journal   = {IEEE Transactions on Biomedical Engineering},
  volume    = {47},
  number    = {9},
  pages     = {1185--1194},
  year      = {2000},
  publisher = {IEEE},
  doi       = {10.1109/10.867928}
}

@article{goldberger2000,
  title     = {{PhysioBank, PhysioToolkit, and PhysioNet}: Components of a New Research Resource for Complex Physiologic Signals},
  author    = {Goldberger, Ary L and Amaral, Luis A N and Glass, Leon and Hausdorff, Jeffrey M and Ivanov, Plamen Ch and Mark, Roger G and Mietus, Joseph E and Moody, George B and Peng, Chung-Kang and Stanley, H Eugene},
  journal   = {Circulation},
  volume    = {101},
  number    = {23},
  pages     = {e215--e220},
  year      = {2000},
  publisher = {American Heart Association},
  doi       = {10.1161/01.CIR.101.23.e215}
}

@article{kadanoff1966,
  title     = {Scaling laws for {Ising} models near $T_c$},
  author    = {Kadanoff, Leo P},
  journal   = {Physics Physique Fizika},
  volume    = {2},
  number    = {6},
  pages     = {263--272},
  year      = {1966},
  publisher = {American Physical Society},
  doi       = {10.1103/PhysicsPhysiqueFizika.2.263}
}

@article{wilsonkogut1974,
  title     = {The renormalization group and the $\epsilon$ expansion},
  author    = {Wilson, Kenneth G and Kogut, John},
  journal   = {Physics Reports},
  volume    = {12},
  number    = {2},
  pages     = {75--199},
  year      = {1974},
  publisher = {Elsevier},
  doi       = {10.1016/0370-1573(74)90023-4}
}

@misc{mehta2014,
  title         = {An exact mapping between the variational renormalization group and deep learning},
  author        = {Mehta, Pankaj and Schwab, David J},
  year          = {2014},
  eprint        = {1410.3831},
  archivePrefix = {arXiv},
  primaryClass  = {cond-mat.dis-nn},
  doi           = {10.48550/arXiv.1410.3831},
  url           = {https://arxiv.org/abs/1410.3831}
}

@article{li2018,
  title     = {Neural Network Renormalization Group},
  author    = {Li, Shuo-Hui and Wang, Lei},
  journal   = {Physical Review Letters},
  volume    = {121},
  number    = {26},
  pages     = {260601},
  year      = {2018},
  publisher = {American Physical Society},
  doi       = {10.1103/PhysRevLett.121.260601}
}

@article{kochjanusz2018,
  title     = {Mutual information, neural networks and the renormalization group},
  author    = {Koch-Janusz, Maciej and Ringel, Zohar},
  journal   = {Nature Physics},
  volume    = {14},
  number    = {6},
  pages     = {578--582},
  year      = {2018},
  publisher = {Nature Publishing Group},
  doi       = {10.1038/s41567-018-0081-4}
}

@article{lin2017,
  title     = {Why does deep and cheap learning work so well?},
  author    = {Lin, Henry W and Tegmark, Max and Rolnick, David},
  journal   = {Journal of Statistical Physics},
  volume    = {168},
  number    = {6},
  pages     = {1223--1247},
  year      = {2017},
  publisher = {Springer},
  doi       = {10.1007/s10955-017-1836-5}
}

@inproceedings{vaswani2017,
  title     = {Attention is all you need},
  author    = {Vaswani, Ashish and Shazeer, Noam and Parmar, Niki and Uszkoreit, Jakob and Jones, Llion and Gomez, Aidan N and Kaiser, {\L}ukasz and Polosukhin, Illia},
  booktitle = {Advances in Neural Information Processing Systems},
  volume    = {30},
  pages     = {5998--6008},
  year      = {2017},
  editor    = {Guyon, I. and Luxburg, U. Von and Bengio, S. and Wallach, H. and Fergus, R. and Vishwanathan, S. and Garnett, R.},
  publisher = {Curran Associates, Inc.},
  url       = {https://proceedings.neurips.cc/paper/2017/file/3f5ee243547dee91fbd053c1c4a845aa-Paper.pdf}
}

\end{document}